%% file: main_ieee.tex
\documentclass[conference]{IEEEtran}
\input{preamble}

\usepackage[hidelinks]{hyperref}
\usepackage{orcidlink}

\begin{document}
\title{ACESO\@: Cross-Modal Distillation for Wearable-Free Freezing of Gait Detection in
Parkinson’s Disease}

\author{\IEEEauthorblockN{Aryan Singh
\orcidlink{0009-0003-0495-627X}}
\IEEEauthorblockA{NeuroAI Fusion Labs\\Kolkata, India\\aryan@neuroailabs.in}
\and
\IEEEauthorblockN{Chandan Biswas
\orcidlink{0000-0003-4468-7396}}
\IEEEauthorblockA{NeuroAI Fusion Labs\\Kolkata, India\\chandan@neuroailabs.in}}

\maketitle

\begin{abstract}
    \input{content/abstract_cb}
\end{abstract}

\begin{IEEEkeywords}
Freezing of Gait, Cross-Modal Distillation, Geometric Self-Occlusion, Knowledge Distillation, Zero-Wearable Inference.
\end{IEEEkeywords}

\section{Introduction}\label{sec:intro}
\input{content/introduction_cb}

\section{Related Work}\label{sec:related_work}
\input{content/related_work_cb}

\section{Background}\label{sec:background}
\input{content/background_cb}

\section{Proposed Approach}\label{sec:proposed}
\input{content/proposed_methods_cb}

\section{Experimental Setup}\label{sec:experimental_setup}
\input{content/exp_setup_cb}

\section{Results and Discussion}\label{sec:results}
\input{content/results_cb}

\section{Conclusion and Future Work}\label{sec:conclusion}
\input{content/conclusion_cb}

\section*{Disclosure}
\input{content/disc}

\bibliographystyle{IEEEtran}
\bibliography{references}
\end{document}

%% file: preamble.tex
\usepackage{booktabs,multirow}
\usepackage{amsmath,amssymb}
\usepackage{graphicx}
\usepackage{tabularx}
\usepackage{array}
\usepackage{colortbl}
\usepackage{microtype}
\usepackage{soul}
\usepackage{xcolor}
\usepackage{float}
\usepackage[ruled,vlined]{algorithm2e}
\usepackage[figureposition=bottom,tableposition=top]{caption}
\newcommand{\para}[1]{\textbf{#1}.}

%% file: content/abstract_cb.tex
Freezing of Gait (FoG) is a paroxysmal disruption of motor skills and control caused by the progressive decline in the basal ganglia's capacity to produce dopamine. FoG manifests suddenly and without warning. Its detection currently relies on body-worn kinematic sensors called inertial measurement units (IMUs). While IMUs provide the best kinematic precision, the need to wear them severely restricts continuous clinical deployment, both logistically and economically. Vision-based detection removes the sensor but is less accurate. To narrow this gap, we propose a supervised cross-modal subspace distillation framework. During training, we turn the kinematic data from the recorded session and the contextual clinical metadata, together with a textual annotation of the video recordings, into oracles that guide a camera-only freeze detector. Empirical evaluations confirm the validity of our approach, which yields 77.3\% balanced accuracy.

%% file: content/introduction_cb.tex
To perform a movement, the body must select one of many possible motor programmes at any given time, much as an operating system's job queue picks a process to run. In humans, this selection is reinforced by the release of dopamine. Parkinson's disease (PD) progressively reduces the brain's ability to produce dopamine. The resulting motor impairment can manifest as an involuntary freeze known as FoG~\cite{nutt2011freezing}, which increases the risk of falls and morbidity~\cite{paul2013three}. The objective detection of FoG events is critical for evaluating disease progression~\cite{moore2008ambulatory} and therapeutic effectiveness~\cite{bachlin2010wearable}.

\begin{figure}[htb]
  \centering
  \includegraphics[width=\linewidth]{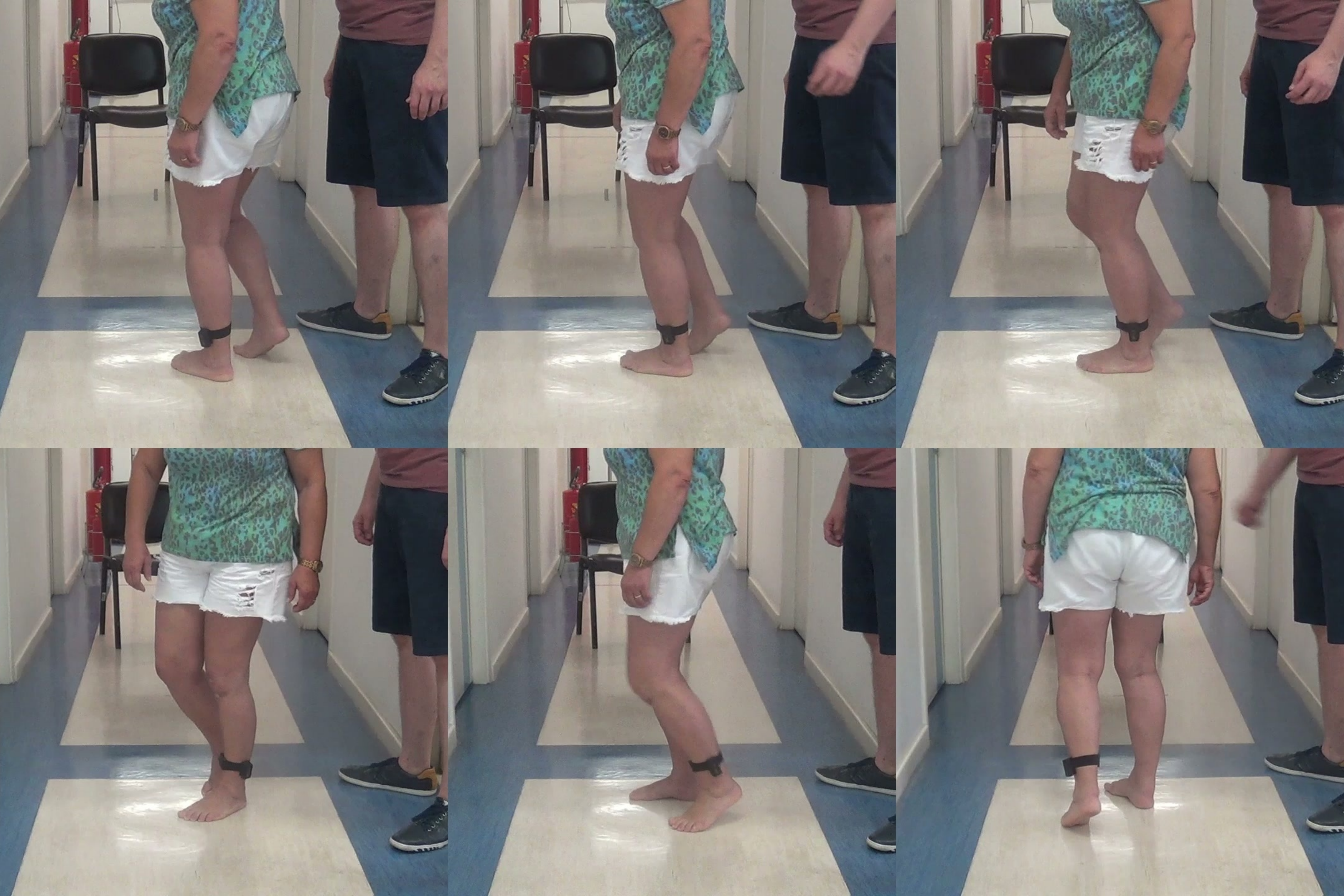}
  \caption{Video frames from a recorded session of the turning-in-place task. The gait narration describes each second of such motion in words: the gait state (e.g.\ turning or shuffling) and attributes such as cadence, step amplitude, foot clearance and direction of travel. The shank-worn IMU is visible on the left leg.}\label{fig:dataset_video}
\end{figure}

PD has no known cure. Our focus is on alleviating the hardship caused by its most debilitating symptom, FoG, whose sudden episodes substantially increase the risk of falls. Named after Aceso (Romanised: \textit{Akeso}), the Greek goddess of the healing process, we propose \textbf{\emph{ACESO}} (Alignment of Clinical and Extrinsic Sensor Observations), a framework for FoG detection that needs no wearable sensor at inference. It uses a spatio-temporal graph convolutional network (ST-GCN)~\cite{yan2018spatial} over all visible body joints. The network is trained as a student of two teachers~\cite{vapnik2015learning}. The first is a kinematic oracle trained on data collected from a body-worn IMU (the extrinsic sensor). The second is a contextual oracle that combines the clinical profile of a subject with a second-by-second description of the video (the clinical observations). The student learns from both teachers by aligning its skeletal embedding with theirs. This embedding is extracted from lower-body video (Figure~\ref{fig:dataset_video}) of subjects performing their task, turning in place. Knowledge distillation~\cite{romero2015fitnets, hinton2015distill} from the oracles to the student embedding happens only during training. The oracles are then dropped, so inference depends on a single video stream.

%% file: content/related_work_cb.tex
Automated FoG detection has progressed from thresholded wearable heuristics to deep inertial models, contactless pose graphs, and cross-modal distillation. Each stage resolved earlier hardware or accuracy limits while exposing the modality gap that our framework addresses.

\para{From Heuristic Wearables to Deep Inertial Models} Early FoG monitors relied on body-worn IMUs, beginning with the spectral Freeze Index, an empirically formulated metric that classified regular gait at $0.5$--$3$\,Hz and FoG at $3$--$8$\,Hz using a shank accelerometer~\cite{moore2008ambulatory}. Subsequent refinements added total-power thresholds to reject stationary standing~\cite{bachlin2010wearable}, knee time-frequency wavelets for subtle treadmill freezes~\cite{delval2010objective}, and deployments using single lumbar or shank sensors~\cite{moore2013autonomous}, smartphones~\cite{capecci2016smartphone}, and real-time cueing~\cite{mazilu2015wearable}. To move beyond rigid thresholds, researchers adopted entropy-based Random Forests~\cite{tripoliti2013automatic}, CNN--LSTM networks over waist spectral windows~\cite{sigcha2020deep}, and a large-scale lower-back sensor contest~\cite{salomon2024machine}. Yet even modern deep TCNs remain vulnerable to task variation, medication-state shifts, voluntary stops~\cite{yang2024freezing}, and poor lab-to-home generalisation. These methods also retain the burden of wearable sensors. Our approach removes test-time wearables entirely, inferring FoG from monocular video while retaining inertial and clinical signals strictly during training.

\para{Rise of Vision-Based Gait and Pose Modelling} To eliminate wearable hardware, vision-based pipelines shifted toward contactless clinical video. Early work quantified MDS-UPDRS gait severity under multi-rater uncertainty~\cite{lu2021quantifying} and detected FoG using anatomical graph-sequence RNNs~\cite{hu2020graph}. Later work advanced to frame-level MS-GCN segmentation on marker-based motion capture~\cite{filtjens2022automated}. More recent video-only methods combine tracking and 3D pose estimation with 1D-CNN--LSTMs for uncontrolled clinical trials~\cite{kondo2024video} or apply Fréchet change-point tests to pose-graph Laplacians~\cite{liu2025graph}. However, monocular keypoints degrade severely under occlusion~\cite{kondo2024video}. Unlike methods that treat all extracted joints equally, our skeleton encoder weights each joint by its tracking confidence to down-weight occluded limbs.

\para{Cross-Modal Distillation and Language Guidance} To recover inertial sensitivity without test-time sensors, generalised distillation~\cite{lopezpaz2016unifying} enabled wearable-to-RGB~\cite{liu2021semantics} and skeleton-to-smartwatch~\cite{ni2022progressive} action recognition. This line of work led to IMU-to-skeleton contrastive distillation for FoG~\cite{tian2025imu2ske}. However, cross-modal transfer can fail when teacher-decisive cues are absent from the student modality~\cite{xue2023modality} or distorted by modality imbalance and soft-label misalignment~\cite{huo2024c2kd}. This helps explain why IMU supervision alone yields moderate sensitivity when turning hides joints~\cite{tian2025imu2ske}. Concurrently, generative action-description prompts~\cite{xiang2023generative} showed that LLM-generated body-part descriptions can guide skeleton encoders during training. Extending both lines of work, we pair confidence-aware visual encoding with shank-IMU distillation and a contextual clinical-text teacher that encodes patient severity and window-level gait dynamics.

\para{Evolution Toward Synchronized Multimodal Benchmarks} Realising this paradigm requires paired data. Earlier unimodal corpora, such as PhysioNet's foot-pressure database ($93$ people with PD and $73$ controls)~\cite{Goldberger2000-xq} and DAPHNet's laboratory IMUs~\cite{bachlin2010wearable}, lacked both video and turning provocation. We therefore evaluate our method on the synchronised benchmark of Ribeiro De Souza et al.~\cite{ribeiro2022public}, which pairs lower-limb video, a shank IMU, and clinical scales for $35$ participants with PD performing turning-in-place tasks. This dataset enables privileged multimodal training with video-only inference.

%% file: content/background_cb.tex
This section reviews the three concepts our method builds on: ST-GCN, knowledge distillation, and Learning Using Privileged Information (LUPI). We first describe our base skeletal network, then explain how distillation transfers knowledge from privileged training modalities into the student skeletal network.

\subsection{Spatial-Temporal Graph Convolutional Networks (ST-GCN)}
\label{subsec:stgcn_background}

ST-GCN~\cite{yan2018spatial} models the human body as an undirected graph $G = (V, E)$. The nodes $V = \{v_{ti} \mid t = 1, \dots, F; i = 1, \dots, K\}$ are the $K$ anatomical joints over a window of $F$ frames, and the edges $E$ connect joints that are physically linked and each joint to itself in the next frame. Given an input $\mathbf{X} \in \mathbb{R}^{C \times F \times K}$ with $C$ channels per joint, ST-GCN alternates spatial and temporal convolutions over this graph.

FoG is characterised by episodic, high-frequency ($3$--$8$\,Hz) festination and a breakdown of coordinated joint displacement. Because ST-GCN reads only the skeleton graph, it ignores appearance and background and captures this breakdown directly, which makes it the encoder of our deployable vision network. Later backbones such as CTR-GCN~\cite{chen2021channel} and InfoGCN~\cite{chi2022infogcn} refine the graph or the embedding, and our distillation objective does not depend on the choice of backbone.

\subsection{Knowledge Distillation}
\label{subsec:kd_background}

Knowledge distillation trains a \emph{student} network to reproduce what a \emph{teacher} network has learnt. Response distillation~\cite{hinton2015distill} softens the teacher's output with a temperature and trains the student to match the resulting distribution, which also conveys the teacher's confidence. Feature distillation~\cite{romero2015fitnets} aligns an intermediate representation of the student with the teacher's through a small regressor (a \emph{hint} head) that bridges the two widths. The first transfers the teacher's decision and the second the structure of its embedding. When the teacher reads a different modality from the student, as in our case, both let the student absorb information it never observes directly.

\subsection{Learning Using Privileged Information (LUPI)}
\label{subsec:lupi_background}

Some modalities, such as wearable inertial sensors and clinical profiles, are available during training but not at telehealth inference. Learning under this condition is known as LUPI~\cite{vapnik2015learning}. Lopez-Paz et al.~\cite{lopezpaz2016unifying} later showed that knowledge distillation is one way to realise LUPI, which is the route we take.

LUPI frameworks use these modalities as expert ``oracles'' that regularise the deployable model. We treat the IMU signal, encoded by a 1D-CNN, and the clinical text, encoded by BioClinicalBERT~\cite{bioclinicalbert}, as such oracles: the vision network is trained to approximate them.

%% file: content/proposed_methods_cb.tex
\begin{figure*}[t]
    \centering
    \includegraphics[width=\textwidth]{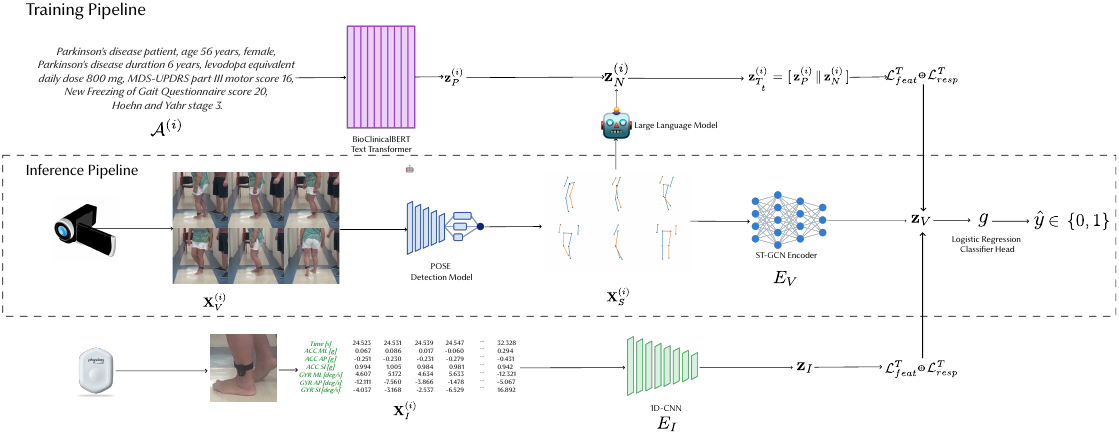}
    \caption{Training and inference pipelines for ACESO\@. During training, a video pose stream, synchronised IMU measurements, and clinical text provide inputs to the visual student and two privileged oracles: a 1D-CNN kinematic encoder and a BioClinicalBERT contextual encoder. The student is trained to detect FoG while distilling knowledge from both oracles. At inference, only the video-derived skeleton is used: the pose sequence is encoded by the ST-GCN, and a logistic regression classifier predicts FoG.}\label{fig:pipeline}
\end{figure*}

This section details the proposed cross-modal distillation framework, ACESO, which allows effective FoG detection from vision-based input alone. Figure~\ref{fig:pipeline} presents a schematic workflow of the method.

During the training phase, an IMU oracle and a clinical contextual oracle are trained or built first and then frozen. A skeleton network is then trained to detect FoG from the video while learning from both oracles. At inference time, the inertial and textual streams are not needed, and classification is executed solely on the visual embedding.

\subsection{Problem Formulation}
Let $\mathcal{D} = \{ (\mathbf{X}_V^{(i)}, \mathbf{X}_I^{(i)}, \mathbf{x}_T^{(i)}, y^{(i)}) \}_{i=1}^N$ be a set of synchronised data instances. For each data instance $i$, $\mathbf{X}_V^{(i)} \in \mathbb{R}^{F \times H \times W \times 3}$ represents a sequence of $F$ visual frames at a spatial resolution of $H \times W$ and $3$ colour channels, from which a pose estimator derives the skeleton sequence $\mathbf{X}_S^{(i)} \in \mathbb{R}^{C \times F \times K}$ with $C$ channels (position, velocity, acceleration and confidence, see Section~\ref{subsec:skeleton_encoder}) for $K$ joints, and $\mathbf{X}_I^{(i)} \in \mathbb{R}^{S \times 6}$ represents the temporally synchronised sequence of $S$ inertial measurements. Additionally, let $\mathbf{x}_T^{(i)} = (\mathbf{x}_P^{(i)}, \mathbf{x}_N^{(i)})$ denote the clinical text of the data instance. It has two parts: the patient profile $\mathbf{x}_P^{(i)}$, a set of sensitive clinical attributes (e.g., age, UPDRS scores) of the recorded session, and the gait narration $\mathbf{x}_N^{(i)}$, a second-by-second description of the movement within the window. The variable $y^{(i)} \in \{0, 1\}$ denotes the ground-truth FoG label (1 if the window overlaps a FoG episode). 

Our objective is to train an encoder on the skeleton sequence $\mathbf{X}_S$ that detects FoG. During this training, the feature set $\mathbf{z}_V$ is distilled with the kinematic (Section~\ref{sec:imu_oracle}) and textual (Section~\ref{sec:text_oracle}) oracles.

\subsection{Kinematic Oracle}
\label{sec:imu_oracle}
Owing to the physical manifestations of FoG episodes, the kinematic data contains a highly discriminative subspace. To leverage it, we use a 1D-CNN:
\begin{equation}
    \mathbf{z}_I = E_I(\mathbf{X}_I; \Theta_I), \quad \mathbf{z}_I \in \mathbb{R}^{p_I}, \qquad p_I = 128
    \label{eq:imu_extractor}
\end{equation}
where $E_I$ is the transformation function that maps the inertial inputs to a $p_I$-dimensional Euclidean space, and $\Theta_I$ denotes the parameters of the IMU encoder, trained together with a linear classification head on the FoG labels and then frozen. The head's logits $\boldsymbol{\ell}_I$ serve as the oracle's prediction.

\subsection{Contextual Oracle}
\label{sec:text_oracle}


The contextual oracle is the concatenation of the patient's clinical profile and a second-by-second narration of the session recording, written by an LLM that is given measurements of the skeletal coordinates. The motivation is to arm the model with the cues a movement expert uses to identify FoG episodes.

\para{Patient profile}
To place the heterogeneous patient metadata in a continuous contextual subspace, we formulate a deterministic prompting strategy. Let $\mathcal{A}^{(i)}$ denote the set of discrete clinical attributes for a given subject instance, specifically incorporating demographic and clinical severity indicators. We define a serialisation function that maps the categorical and numerical variables of $\mathcal{A}^{(i)}$ into one natural language sentence, $\mathbf{x}_P^{(i)}$.
The sentence is processed by the frozen BioClinicalBERT~\cite{bioclinicalbert} transformer, which produces $\mathbf{z}_P^{(i)} \in \mathbb{R}^{768}$.

\para{Gait narration}
The narration is computed from the same skeleton the vision network sees, never from the FoG annotations. For each second, we measure lower-limb motion in leg lengths, so that camera distance cancels: cadence, step amplitude, foot clearance, foot travel and its direction, turning rate, and the freeze index~\cite{moore2008ambulatory,bachlin2010wearable}. We describe these numbers with a closed vocabulary of $47$ words: one of $8$ gait states, one value for each of $7$ attributes (cadence, amplitude, clearance, progression, oscillation, rhythm, transition), and $1$ out of $10$ travel directions.

We study two narration variants, which differ only in who assigns the words. In the \emph{LLM} variant, a large language model (Claude Sonnet 5.5) receives the measurements of a whole session in a single prompt, together with a definition of every state and its thresholds, and returns one record per second. In the \emph{rules} (programmatic) variant, a fixed cascade of thresholds on the same measurements assigns the words, with no language model. We use only the LLM variant for our proposed method. Both variants yield the narration vector $\mathbf{z}_N^{(i)} \in \mathbb{R}^{47}$, which holds, for each of the $47$ words, the share of the window's seconds that carry it, so that each word group sums to one.

The final text vector $\mathbf{z}_T^{(i)}$ is the concatenation of the two blocks. It is built once for every window, before any training and independently of the folds, and serves as the contextual oracle during distillation:

\begin{equation}
\mathbf{z}_T^{(i)} = [\,\mathbf{z}_P^{(i)} \,\|\, \mathbf{z}_N^{(i)}\,] \in \mathbb{R}^{815}
\end{equation}

A linear classification head $f_T$ is trained on $\mathbf{z}_T$ with the FoG labels and then frozen. Its logits serve as the oracle's prediction:
\begin{equation}
    \boldsymbol{\ell}_T = f_T(\mathbf{z}_T) = \mathbf{W}_T\, \mathbf{z}_T + \mathbf{c}_T, \qquad \boldsymbol{\ell}_T \in \mathbb{R}^{2}
    \label{eq:text_head}
\end{equation}
where $\mathbf{W}_T \in \mathbb{R}^{2 \times 815}$ and $\mathbf{c}_T \in \mathbb{R}^{2}$ are the weights and the bias of the head.

\subsection{Confidence-Gated ST-GCN}
\label{subsec:skeleton_encoder}

We do not have a theoretical framework that explains when and how a FoG episode is triggered, but empirical evidence establishes that sharp turns can easily trigger one. To study FoG, the dataset authors therefore had the subjects perform a turning-in-place task at their own pace~\cite{ribeiro2022public}. This task produces skeleton graphs that are often occluded, with joints lost or tracked with low confidence. We do not repair these joints. Instead, the network learns how much to trust each one.

A pose estimator~\cite{jocher_yolo11} with a tracker~\cite{Aharon2022BoTSORTRA} identifies $K$ anatomical joints in each frame, with image coordinates and confidence $c_j^t \in [0,1]$. The skeleton input $\mathbf{X}_S \in \mathbb{R}^{C \times F \times K}$ has $C = 7$ channels for each joint: two for position, two for velocity, two for acceleration, and one for confidence.

The encoder $E_V$ is an ST-GCN~\cite{yan2018spatial} of three graph blocks $b \in \{1,2,3\}$ with $64$, $128$ and $256$ channels. Before each block $b$, the features of joint $j$ at frame $t$ are multiplied by a weight:
\begin{equation}
    w_{b,j}^{t} = \sigma\big(s_b \, c_j^{t} + o_b\big)
    \label{eq:conf_gate}
\end{equation}
where $c_j^t$ is the joint's pose confidence, $\sigma$ is the sigmoid function, and $s_b$ (scale) and $o_b$ (offset) are two numbers learnt for each block.

Each block then mixes the weighted features across neighbouring joints with a graph convolution over the skeleton, and across neighbouring frames with a temporal convolution. The output of the third block is averaged over all frames and all joints, which gives one visual embedding for each window:
\begin{equation}
    \mathbf{z}_V = E_V(\mathbf{X}_S; \Theta_V) \in \mathbb{R}^{p}, \qquad p = 256
    \label{eq:visual_emb}
\end{equation}
where $\Theta_V$ holds the weights of the three blocks together with the gate parameters $\{s_b, o_b\}_{b=1}^{3}$, and $p$ is the width of the last block.

\subsection{Supervised Cross-Modal Distillation}\label{subsec:distillation}


During training, the student ($\mathbf{z}_V$) learns from the ground-truth FoG labels and from two privileged teachers. We transfer the teachers' knowledge in two complementary ways. Response distillation~\cite{hinton2015distill} teaches the student to match each teacher's FoG predictions, including their relative confidence. Feature distillation~\cite{romero2015fitnets} teaches it to match useful internal representations learnt by the teachers. Together, they guide both the student's decisions and the information it learns to represent. The following subsections describe each approach and its loss function in detail.

\para{Response distillation} 
Each oracle $m \in \{I,T\}$ has a frozen head with logits $\boldsymbol{\ell}_m$. The student has a linear classification head that produces the logits $\boldsymbol{\ell}_V$ from $\mathbf{z}_V$. The student matches each oracle's temperature-softened class predictions using the KL divergence from the oracle of $\boldsymbol{\ell}_m$ to the student of $\boldsymbol{\ell}_V$, denoted $\mathcal{L}_{resp}^{m}$, where $\tau$ is the distillation temperature:

\begin{equation}
    \mathcal{L}_{resp}^{m} = \tau^{2}\, \mathrm{KL}\Big( \mathrm{softmax}\big(\boldsymbol{\ell}_m / \tau\big) \,\Big\|\, \mathrm{softmax}\big(\boldsymbol{\ell}_V / \tau\big) \Big)
    \label{eq:response_kd}
\end{equation}

\para{Feature distillation}
The profile block $\mathbf{z}_P$ of the contextual oracle vector is the same for a patient's FoG and non-FoG windows, so $\mathbf{z}_T$ alone is a weak target for telling the two classes apart. We therefore replace it, as a training target only, with a class-conditioned vector $\tilde{\mathbf{z}}_T$, obtained from a per-class affine projector:
\begin{equation}
    \tilde{\mathbf{z}}_T^{(i)} = \mathbf{z}_T^{(i)} \odot (\mathbf{1} + \mathbf{a}_{y^{(i)}}) + \mathbf{b}_{y^{(i)}}
    \label{eq:text_projector}
\end{equation}
where $y^{(i)}$ is the FoG label of window $i$ and $\mathbf{a}_y,\mathbf{b}_y \in \mathbb{R}^{d_T}$, with $d_T = 815$, are fitted per class on training windows with cross-entropy over cosine class prototypes and an $\ell_2$ penalty towards the identity, and then frozen.

Two linear hint heads, $h_I: \mathbb{R}^{p} \rightarrow \mathbb{R}^{p_I}$ and $h_T: \mathbb{R}^{p} \rightarrow \mathbb{R}^{d_T}$, map $\mathbf{z}_V$ to each oracle's width, and the mapped embeddings are aligned with the oracle targets using cosine distance:
\begin{equation}
    \begin{aligned}
    \mathcal{L}_{feat}^{I} &= 1 - \cos\big(h_I(\mathbf{z}_V), \mathbf{z}_I\big), \\
    \mathcal{L}_{feat}^{T} &= 1 - \cos\big(h_T(\mathbf{z}_V), \tilde{\mathbf{z}}_T\big)
    \end{aligned}
    \label{eq:feature_kd}
\end{equation}

The student is trained with a class-weighted cross-entropy on the FoG labels together with the four distillation terms:
\begin{equation}
    J = \mathcal{L}_{CE}\big(\boldsymbol{\ell}_V, y\big) + \sum_{m \in \{I, T\}} \Big( \lambda_m \, \mathcal{L}_{resp}^{m} + \gamma_m \, \mathcal{L}_{feat}^{m} \Big)
    \label{eq:total_loss}
\end{equation}

In $\mathcal{L}_{CE}$, a FoG window is weighted by $N_0 / N_1$ and a non-FoG window by one, where $N_0$ and $N_1$ are the numbers of non-FoG and FoG windows in the training set, so that both classes contribute equally to the loss. The loss $J$ is minimised over the encoder weights $\Theta_V$, the student's classification head, and the two hint heads $h_I$ and $h_T$. The scalars $\lambda_m \geq 0$ and $\gamma_m \geq 0$ are fixed hyperparameters, one pair for each oracle $m \in \{I, T\}$. They are set before training and are not learnt. $\lambda_m$ scales the response loss $\mathcal{L}_{resp}^{m}$, which controls how strongly the student is pulled towards the class predictions of oracle $m$. $\gamma_m$ scales the feature loss $\mathcal{L}_{feat}^{m}$, which controls how strongly the student embedding is pulled towards the embedding of oracle $m$. The cross-entropy term has a fixed weight of one, so each $\lambda_m$ and $\gamma_m$ is read relative to the supervised loss. 

\begin{equation}
    \hat{y} = g(\mathbf{z}_V) \in \{0, 1\}
    \label{eq:prediction}
\end{equation}

After training, the encoder $E_V$ is frozen, and its embedding $\mathbf{z}_V$ (Eq.~\eqref{eq:visual_emb}) is the feature set of each window. A separate classifier $g$ is then fitted on $\mathbf{z}_V$ to label each window as FoG or non-FoG (Section~\ref{subsec:classifier_robustness}). By default, $g$ is logistic regression (LR), which produces the prediction $\hat{y}$. 

The detailed steps are given in Algorithm~\ref{alg:distillation}.

\input{content/algo_cb}

%% file: content/algo_cb.tex
\begin{algorithm}[h!tbp]
\small
\DontPrintSemicolon
\caption{Supervised Cross-Modal Distillation}
\label{alg:distillation}

\KwIn{
  Training windows $\mathcal{D}_{tr} = \{(\mathbf{X}_S^{(i)}, \mathbf{X}_I^{(i)}, \mathbf{x}_T^{(i)}, y^{(i)})\}_{i=1}^N$, with $\mathbf{x}_T^{(i)} = (\mathbf{x}_P^{(i)}, \mathbf{x}_N^{(i)})$ \\
  Batch size $M$, temperature $\tau$, weights $\lambda_I, \lambda_T, \gamma_I, \gamma_T$
}
\KwOut{
  Encoder weights $\Theta_V$ and classifier $g$
}

\BlankLine
Train the IMU encoder $E_I(\cdot\,;\Theta_I)$ and its linear head on $(\mathbf{X}_I, y)$ (Eq.~\ref{eq:imu_extractor}), then freeze both\
\ForEach{window $i \in \mathcal{D}_{tr}$}{
    $\mathbf{z}_P^{(i)} \gets \text{BioClinicalBERT}(\mathbf{x}_P^{(i)})$\;
    $\mathbf{z}_N^{(i)} \gets$ word shares of $\mathbf{x}_N^{(i)}$\;
    $\mathbf{z}_T^{(i)} \gets [\,\mathbf{z}_P^{(i)} \,\|\, \mathbf{z}_N^{(i)}\,]$\;
}
Train the text head $f_T$ on $(\mathbf{z}_T, y)$ (Eq.~\ref{eq:text_head}), then freeze it\;
Fit the projector parameters $\{\mathbf{a}_y, \mathbf{b}_y\}_{y \in \{0,1\}}$ on $(\mathbf{z}_T, y)$, then freeze them\;

\BlankLine
Initialise $\Theta_V$, the student's classification head and the hint heads $h_I, h_T$\;
\Repeat{the detection AUC on the inner selection subjects stops improving}{
    \ForEach{minibatch $\mathcal{B} \subset \mathcal{D}_{tr}$ such that $|\mathcal{B}| = M$}{
        \For{$i \in \mathcal{B}$}{
            $\mathbf{z}_I^{(i)} \gets E_I(\mathbf{X}_I^{(i)}; \Theta_I)$, \quad $\boldsymbol{\ell}_I^{(i)} \gets$ IMU head$(\mathbf{z}_I^{(i)})$\;
            $\boldsymbol{\ell}_T^{(i)} \gets f_T(\mathbf{z}_T^{(i)})$\;
            $\tilde{\mathbf{z}}_T^{(i)} \gets \mathbf{z}_T^{(i)} \odot (\mathbf{1} + \mathbf{a}_{y^{(i)}}) + \mathbf{b}_{y^{(i)}}$ (Eq.~\ref{eq:text_projector})\;
            $\mathbf{z}_V^{(i)} \gets E_V(\mathbf{X}_S^{(i)}; \Theta_V)$ (Eq.~\ref{eq:visual_emb}), \quad $\boldsymbol{\ell}_V^{(i)} \gets$ classification head$(\mathbf{z}_V^{(i)})$\;
        }
        Compute $\mathcal{L}_{resp}^{m}$ (Eq.~\ref{eq:response_kd}) and $\mathcal{L}_{feat}^{m}$ (Eq.~\ref{eq:feature_kd}) for $m \in \{I, T\}$, averaged over $\mathcal{B}$\;
        $J \gets \mathcal{L}_{CE}\big(\boldsymbol{\ell}_V, y\big) + \sum_{m \in \{I,T\}} \big(\lambda_m \mathcal{L}_{resp}^{m} + \gamma_m \mathcal{L}_{feat}^{m}\big)$ (Eq.~\ref{eq:total_loss})\;
        Update $\Theta_V$, the classification head and $h_I, h_T$ by a gradient step on $J$\;
    }
}

\BlankLine
Freeze $E_V$ with $\Theta_V$ from the epoch with the best selection AUC\;
\ForEach{window $i \in \mathcal{D}_{tr}$}{
    $\mathbf{z}_V^{(i)} \gets E_V(\mathbf{X}_S^{(i)}; \Theta_V)$\;
}
Fit the classifier $g$ on $(\mathbf{z}_V, y)$ (Eq.~\ref{eq:prediction})\;
\Return $\Theta_V$, $g$\;
\end{algorithm}

%% file: content/exp_setup_cb.tex

This section summarises the dataset and details the cross-validation protocol, the implementation, the experimental configurations, and the evaluation metrics.

\subsection{Dataset and Partitioning Protocol}\label{subsec:dataset_protocol}

We use the public FoG dataset~\cite{ribeiro2022public}, which includes $35$ participants with PD who perform a two-minute turning-in-place task. The dataset includes lower-limb video recorded at $30$\,Hz, a shank-worn IMU recorded at $128$\,Hz on the more affected leg~\cite{ribeiro2022public}, and FoG annotations from two movement experts. Each subject was scheduled for three sessions, but the dataset contains an average of $2.1$ sessions per subject. Figure~\ref{fig:dataset_video} shows example frames from a recorded session. The dataset is partitioned into five subject-disjoint folds so that all sessions from a participant fall in the same fold. We train the model on four folds and test it on the fifth, repeating this process until each fold has served as the test fold once. Each fold contains about $1650$ windows on average, of which $31.4\%$ are positive for FoG (maximum $41.8\%$, minimum $16.9\%$).

\subsection{Implementation Details}\label{subsec:implementation_details}

Windows are $7.8125$\,s long ($F = 234$ frames, $1000$ IMU samples) with a $1$\,s stride. The IMU oracle is a 1D-CNN with channels $64$, $64$ and $128$ and kernels $7$, $5$ and $3$. Poses come from YOLO11x-pose with BoT-SORT. The ST-GCN uses a temporal kernel of $9$. The student is trained on Eq.~(\ref{eq:total_loss}) with $M = 128$, $\tau = 4$, $\lambda_I = \gamma_I = 1$ and $\lambda_T = \gamma_T = 0.5$, using Adam ($\eta = 5 \times 10^{-4}$, weight decay $10^{-4}$, cosine decay) and early stopping (patience $200$ epochs). BA and sensitivity depend on the $\theta$ applied to $g$. Within each fold, $\theta$ was selected to maximise $J$ on the training folds of the training data, then applied unchanged to the testing fold. This procedure was identical across configurations and used no test windows for threshold selection. The mean threshold across five folds was $0.34$ for ACESO and $0.23$ for the un-distilled baseline.

\subsection{Baseline and Ablation Groups}\label{subsec:baselines}

Each configuration uses either a frozen embedding or a concatenation of embeddings, classified with an LR head\@.

\noindent\textbf{Un-distilled}: the same ST-GCN Eq.~(\ref{eq:total_loss} with $\lambda_m = \gamma_m = 0$ for both oracles). This and ACESO differ only in the distillation terms. \\
\noindent\textbf{ACESO}: the ST-GCN features distilled from the kinematic and contextual oracles (Eq.~(\ref{eq:total_loss})), which are dropped after training. \\
\noindent\textbf{Kinematic Oracle} ($\mathbf{z}_I$): the IMU oracle, scored as an upper bound based on a body-worn sensor. \\
\noindent\textbf{Contextual Oracle} ($\mathbf{z}_T$): the LLM-based or rules-based contextual vector. \\
\noindent\textbf{Feature Concatenation} ($\mathbf{z}_A \oplus \mathbf{z}_B$): 
To validate our hypothesis, we report comparable results across concatenated modalities, in which two or more frozen feature embeddings are concatenated before the classification head. We evaluate ACESO alone and concatenated with the IMU features, the clinical-text features, or both. We also evaluate the IMU and text features together, and the LLM and rules narrations alone and alongside the student. All configurations use the same folds and the same classifier procedure.


\subsection{Evaluation Metrics}
\label{subsec:eval_metrics}

\begin{figure}[htbp]
  \centering
  \includegraphics[width=\linewidth]{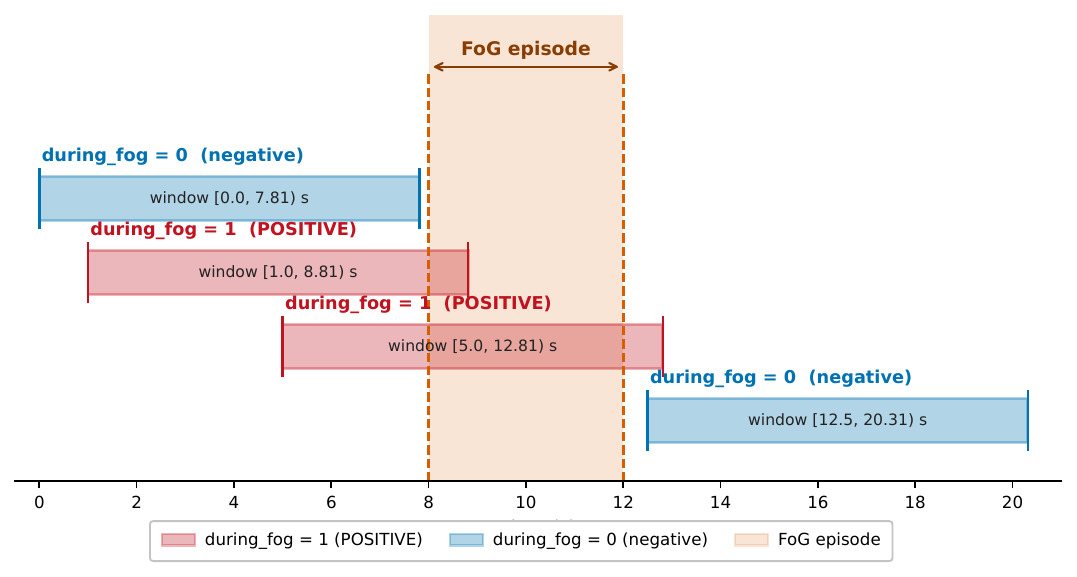}
  \caption{Strategy used to label windows.}\label{fig:label}
\end{figure}

Since only $31.4\%$ of windows overlap a FoG episode, we report balanced accuracy (BA), $\mathrm{BA}=(\mathrm{Sensitivity}+\mathrm{Specificity})/2$, and ROC-AUC\@. As shown in Figure~\ref{fig:label}, a window is positive if it overlaps an annotated FoG episode. We also report the median detection latency across detected episodes:
\begin{equation}
  \Delta t = t_{\text{alarm}} - t_{\text{onset}}, \qquad
  t_{\text{alarm}}=\text{end of first positive window}.
\end{equation}
Here, $t_{\text{onset}}$ is the annotated episode onset. Using the window end reflects that the full window is required to make a prediction.

%% file: content/results_cb.tex
The experiments evaluate whether privileged supervision from inertial and clinical signals improves vision-only FoG detection, examined in turn through ablation, comparison with existing frameworks, and sensitivity to the choice of classifier.

\subsection{Overall Detection Performance and Ablation}\label{subsec:overall_performance}

Our hypothesis is that distilling kinematic and contextual oracles into the skeletal features of a video improves the representation even when the oracles are absent at inference. To evaluate this, Table~\ref{tab:performance} reports the detection performance of ACESO and of each ablation group.

\begin{table}[htbp]
\caption{Overall performance of isolated and concatenated feature configurations. All metrics use the LR head. (LLM) and (rules) mark the source of the Contextual Oracle $\mathbf{z}_T$.}\label{tab:performance}
\renewcommand{\arraystretch}{1.2}
\resizebox{\columnwidth}{!}{%
\begin{tabular}{l c c c}
\toprule
\textbf{Feature Configuration} & \textbf{BA} & \textbf{ROC-AUC} & \textbf{Sensitivity} \\
\midrule
\multicolumn{4}{c}{\textit{Vision-Only Inference}} \\
Un-distilled & 73.6\% & 84.1\% & 73.0\% \\
\textbf{ACESO} & \textbf{77.3\%} & 84.0\% & 71.4\% \\
\midrule
\multicolumn{4}{c}{\textit{Multi-Modal Inference}} \\
Un-distilled $\oplus \mathbf{z}_I$ & 74.6\% & 87.2\% & \textbf{80.5\%} \\
ACESO $\oplus \mathbf{z}_I$ & 74.1\% & 87.1\% & 75.1\% \\
Un-distilled $\oplus \mathbf{z}_T$ (LLM) & 63.8\% & 70.2\% & 59.5\% \\
ACESO $\oplus \mathbf{z}_T$ (LLM) & 73.0\% & 81.8\% & 75.3\% \\
Un-distilled $\oplus \mathbf{z}_T$ (rules) & 66.7\% & 78.0\% & 69.3\% \\
ACESO $\oplus \mathbf{z}_T$ (rules) & 65.7\% & 73.9\% & 67.2\% \\
ACESO $\oplus \mathbf{z}_N$ (LLM) & 76.1\% & 83.4\% & 70.2\% \\
ACESO $\oplus \mathbf{z}_N$ (rules) & 74.6\% & 80.4\% & 70.4\% \\
$\mathbf{z}_I \oplus \mathbf{z}_T$ (LLM) & 72.4\% & 87.3\% & 76.3\% \\
$\mathbf{z}_I \oplus \mathbf{z}_T$ (rules) & 72.6\% & 86.1\% & 77.4\% \\
Un-distilled $\oplus \mathbf{z}_I \oplus \mathbf{z}_T$ (LLM) & 72.4\% & \textbf{88.4\%} & 75.8\% \\
ACESO $\oplus \mathbf{z}_I \oplus \mathbf{z}_T$ (LLM) & 73.3\% & 86.1\% & 75.2\% \\
Un-distilled $\oplus \mathbf{z}_I \oplus \mathbf{z}_T$ (rules) & 72.5\% & 87.4\% & 79.2\% \\
ACESO $\oplus \mathbf{z}_I \oplus \mathbf{z}_T$ (rules) & 72.1\% & 84.3\% & 71.4\% \\
\midrule
\multicolumn{4}{c}{\textit{Privileged Oracles}} \\
Kinematic Oracle & 74.2\% & 89.6\% & 78.8\% \\
Contextual Oracle (LLM) & 49.4\% & 49.3\% & 53.6\% \\
Contextual Oracle (rules) & 53.6\% & 51.9\% & 32.6\% \\
Gait narration $\mathbf{z}_N$ (LLM) & 73.9\% & 83.6\% & 69.2\% \\
Gait narration $\mathbf{z}_N$ (rules) & 74.9\% & 82.4\% & 76.8\% \\
\bottomrule
\end{tabular}
}
\end{table}

\begin{figure}[htb]
  \centering
  \includegraphics[width=\linewidth]{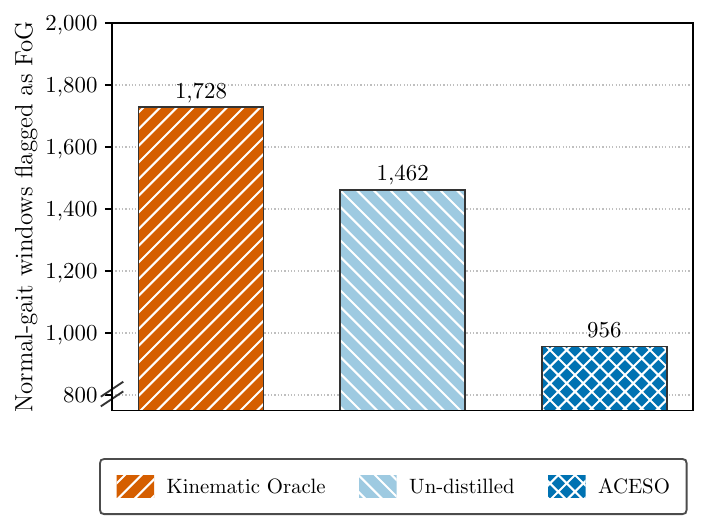}
  \caption{Number of false-positive windows for each configuration. The vertical axis does not start at zero.}\label{fig:false_alarms}
\end{figure}

Table~\ref{tab:performance} supports this hypothesis. Distillation raises BA by $3.7\%$ points. Sensitivity falls by $1.6\%$ points, but specificity rises from $74.2\%$ to $83.2\%$, which means ACESO raises fewer false alarms than the un-distilled baseline (Figure~\ref{fig:false_alarms}). Among the multi-modal configurations, the highest ROC-AUC comes from Un-distilled $\oplus \mathbf{z}_I \oplus \mathbf{z}_T$ (LLM), which carries the most information and therefore separates freezing from non-freezing windows best.

Since BA depends on the threshold, we also report threshold-free and episode-level results for the LR head. AP is $78.1\%$ for ACESO, $77.0\%$ for the un-distilled model and $82.3\%$ for the kinematic oracle. At the episode level, ACESO detects $127$ of $173$ episodes ($73.4\%$) with $1.67$ false alarms per minute, against $145$ episodes ($83.8\%$) and $1.76$ false alarms per minute for the un-distilled model. The gain of distillation is therefore in window-level specificity, not in the number of episodes detected.

When both oracles are concatenated to the vision features, the un-distilled baseline attains a higher ROC-AUC than ACESO, by $2.3\%$ points with the LLM contextual oracle and by $3.1$ with the rules-based one. Adding the IMU and LLM features does not consistently improve performance. The un-distilled combination achieves the highest ROC-AUC ($88.4\%$), but ACESO performs better in BA ($73.3\%$ vs.\ $72.4\%$), and the rules-based combinations score lower in BA than vision-only ACESO\@.

\begin{figure}[htbp]
  \centering
  \includegraphics[width=\linewidth]{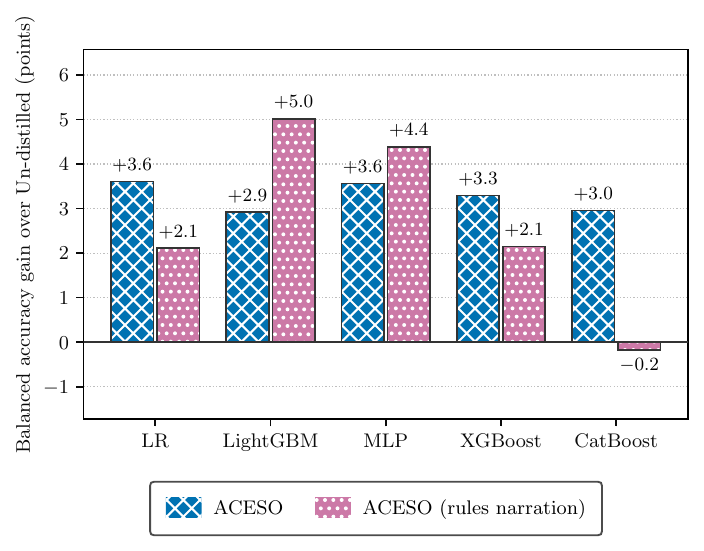}
  \caption{Gain in BA (percentage points) of the distilled vision-only model over the un-distilled model, per classification head. ACESO is trained with the LLM narration, and ACESO (rules narration) with the rules narration.}
  \label{fig:camera_heads}
\end{figure}

Figure~\ref{fig:camera_heads} shows that distillation improves BA over the un-distilled model for nine of the ten classifier and narration settings. The largest gain is $5.0\%$ points for LightGBM with the rules narration, while the only loss is $0.2\%$ points for CatBoost with the rules narration. The rules narration gives a higher BA than the LLM narration for LightGBM and the MLP, while the LLM narration performs better for the other three heads. We retain the LLM narration because LR was selected as the primary head before this comparison, and the LLM narration leads by $1.5\%$ points with LR.

\begin{figure*}[!t]
  \centering
  \includegraphics[width=\textwidth]{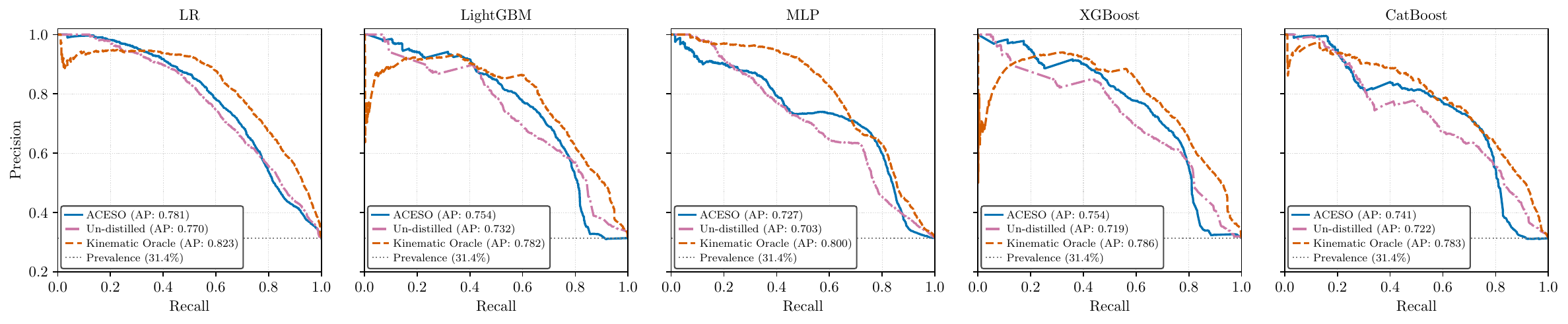}
  \caption{PR curves of ACESO, the un-distilled baseline and the privileged kinematic oracle under each classification head, with AP\@. The dashed line marks the positive-class prevalence.}

  \label{fig:pr}
\end{figure*}

\begin{figure}[htbp]
  \centering
  \includegraphics[width=0.8\linewidth]{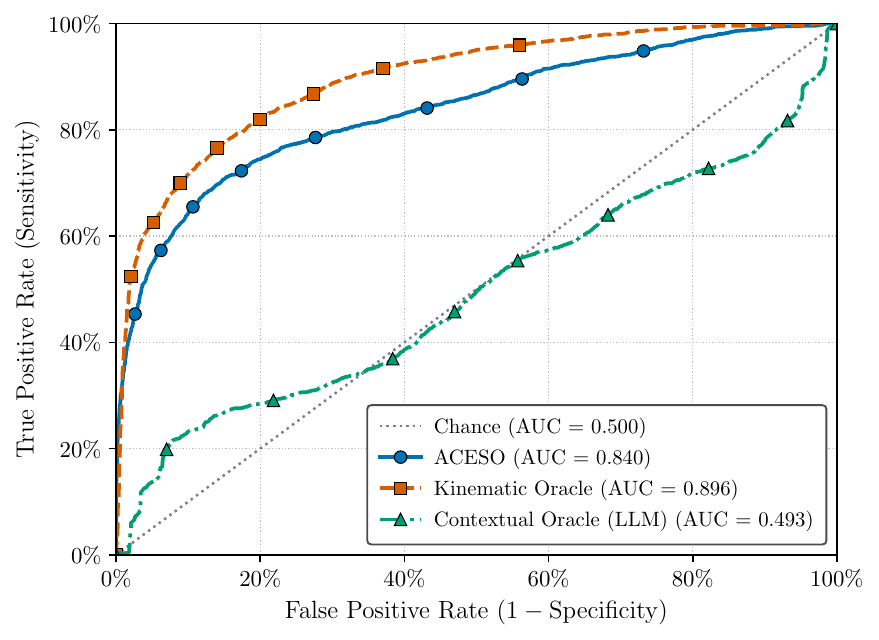}
  \caption{ROC curves evaluating the deployable vision-only configuration (ACESO) against the privileged train-only oracles ($\mathbf{z}_I$ and $\mathbf{z}_T$) under the identical LR classification head.}
  \label{fig:roc}
\end{figure}

Figure~\ref{fig:roc} shows the ROC curves. Using only visual features at inference, ACESO yields a ROC-AUC $5.6$ percentage points below that of the kinematic oracle. The full contextual oracle (LLM) is at chance level on its own ($49.3\%$), whereas its gait narration alone reaches a ROC-AUC of $83.6\%$ (Table~\ref{tab:performance}). The narration therefore carries the detection signal for the contextual oracle. The visual student recovers most, but not all, of the discriminative information in the inertial modality without requiring IMU measurements at deployment.

\subsection{Comparison with Competing Frameworks}\label{subsec:sota_comparison}
Prior work on FoG analysis has relied predominantly on wearable inertial sensing or on small cohorts, and differs from ours in modality, task formulation and evaluation protocol. Table~\ref{tab:sota-comparison} situates ACESO, which requires only video at inference, among these approaches. All rows are as reported by their authors, with their own windows, label definitions and splits, so the table gives context and not a controlled comparison. IMU2SKE~\cite{tian2025imu2ske} uses the same 35-subject dataset.
%
\begin{table}[hbp]
\caption{Comparison with SOTA methods for FoG analysis.}\label{tab:sota-comparison}
\renewcommand{\arraystretch}{1.2} 
\resizebox{\columnwidth}{!}{
\begin{tabular}{@{}l l l c c c c@{}}
\toprule
\textbf{Study} & \textbf{Modality} & \textbf{Task} & \textbf{Horiz./Latency} & \textbf{Subjects} & \textbf{BA} & \textbf{Sensitivity} \\
\midrule
Zhang (2020)~\cite{zhang2020prediction} & IMU & Pred. & 0.93 s & 12 & 75.8 & 72.7 \\
Huang (2024)~\cite{huang2024episode} & IMU & Pred. & 5.0 s & 12 & 79.2 & 72.9 \\
Kondo (2024)~\cite{kondo2024video} & Vision$^{\dagger}$ & Det. & N/A & 16$^{\ddagger}$ & 93.0 & 88.0 \\
Tian (2025)~\cite{tian2025imu2ske} & Vision & Det. & 3.0 s & 35 & 70.1 & 51.1 \\
\midrule
\textbf{ACESO} & \textbf{Vision} & \textbf{Det.} & \textbf{0.86 s (Latency)} & \textbf{35} & \textbf{77.3} & \textbf{71.4} \\
\bottomrule
\end{tabular}%
}
\vspace{0.1cm}
\raggedright
\scriptsize{$^{\dagger}$ Uses an expert-guided CNN-LSTM and manual preprocessing. \\
$^{\ddagger}$ Includes 9 PD and 7 Progressive Supranuclear Palsy subjects. All other studies were evaluated exclusively on PD.}
\end{table}
Operating on vision alone at inference, ACESO yields $77.3\%$ BA, $71.4\%$ sensitivity, and $83.2\%$ specificity. On the same 35-subject dataset, IMU2SKE~\cite{tian2025imu2ske} scores $70.1\%$ BA, $51.1\%$ sensitivity, and $89.1\%$ specificity. It raises its first alarm a median of $0.86$\,s after the freeze begins, and $59\%$ of detected episodes are flagged within the first second.

\subsection{Competing Classification Algorithms}\label{subsec:classifier_robustness}

The primary classification head of ACESO is LR\@. To establish that its performance does not depend on this choice, we repeat the experiments with different classifiers. In Eq.~\eqref{eq:prediction}, $g$ can be replaced by other functions. We use a multi-layer perceptron (MLP) optimised with a class-weighted objective and three gradient-boosted tree ensembles (LightGBM, XGBoost, and CatBoost), with all other parameters fixed. Table~\ref{tab:aceso_heads} reports the results.

\begin{table}[htbp]
\caption{Performance stability of ACESO evaluated across varying linear and non-linear classification heads.}
\label{tab:aceso_heads}
\centering
\small
\begin{tabular}{l c c c}
\toprule
\textbf{Classification Head} & \textbf{BA} & \textbf{ROC-AUC} & \textbf{Sensitivity} \\
\midrule
LightGBM & 76.7\% & 79.3\% & 72.5\% \\
MLP & \textbf{78.5\%} & 81.2\% & \textbf{77.2\%} \\
XGBoost & 77.2\% & 80.2\% & 73.6\% \\
CatBoost & 77.7\% & 79.5\% & 71.0\% \\
LR & 77.3\% & \textbf{84.0\%} & 71.4\% \\
\bottomrule
\end{tabular}
\end{table}

Table~\ref{tab:aceso_heads} shows that BA varies by less than $2\%$ points across the classifiers ($76.7\%$ to $78.5\%$). ROC-AUC varies more, from $79.3\%$ to $84.0\%$, so the ranking of windows depends on the head while the balance at the operating point does not. Because FoG windows are the minority class, precision-recall (PR) analysis provides a stricter assessment than ROC, with positive-class prevalence acting as a baseline (Figure~\ref{fig:pr}). ACESO exceeds this reference under every head and achieves higher AP than the undistilled model, with gains ranging from $1.1\%$ points for LR to $3.5\%$ points for XGBoost. ACESO's AP advantage is concentrated at low-to-moderate recall. Beyond a recall of approximately $0.8$, the undistilled model achieves higher precision. Since the kinematic oracle maintains higher precision across all heads, we infer that the remaining gap is associated with the inertial input rather than the classifier choice. LR achieves the highest AP for ACESO, consistent with its ROC-AUC performance, and is therefore used as the default classification head $g$.



%% file: content/conclusion_cb.tex
Cross-modal distillation improves the balance of vision-only FoG detection, showing that inertial and clinical context can shape a useful skeleton representation even when those signals are absent at deployment. The gain is primarily in the operating point, with little change in window ranking. The encoder takes pose confidence as an input and a gate, but we did not isolate its effect on occluded windows.

The method requires synchronised multimodal data for training, and its evaluation on one cohort does not establish broader generalisation or FoG anticipation. Future work should validate across independent populations and explore temporal prediction with incomplete or asynchronous training signals.

%% file: content/disc.tex
Generative AI models such as Anthropic's Claude (Sonnet and Opus 5.5, 4.8 and 4.7) and Google's Gemini were employed exclusively for code implementation, code review, and grammar and spell checking.